# Combining Context and Knowledge Representations for Chemical-disease Relation Extraction

Huiwei Zhou, Yunlong Yang, Shixian Ning, Zhuang Liu, Chengkun Lang, Yingyu Lin, Degen Huang

*Abstract*—Automatically extracting the relationships between chemicals and diseases is significantly important to various areas of biomedical research and health care. Biomedical experts have built many large-scale knowledge bases (KBs) to advance the development of biomedical research. KBs contain huge amounts of structured information about entities and relationships, therefore plays a pivotal role in chemical-disease relation (CDR) extraction. However, previous researches pay less attention to the prior knowledge existing in KBs. This paper proposes a neural network-based attention model (NAM) for CDR extraction, which makes full use of context information in documents and prior knowledge in KBs. For a pair of entities in a document, an attention mechanism is employed to select important context words with respect to the relation representations learned from KBs. Experiments on the BioCreative V CDR dataset show that combining context and knowledge representations through the attention mechanism, could significantly improve the CDR extraction performance while achieve comparable results with state-of-the-art systems.

## I. INTRODUCTION

Extracting semantic relations between chemicals and diseases in the biomedical literature is one of the main tasks in the precision medical treatment. It is of essential importance to the clinical disease diagnosis, treatment and drug development [1], [2]. However, manually extracting these relations from the biomedical literature into structured knowledge bases, such as Comparative Toxicogenomics Database (CTD) [3], is expensive and time-consuming, and it is difficult to keep up-to-date.

To promote research on these issues, the BioCreative-V community [4] proposes a challenging task of automatically extracting CDR from biomedical literature. It consists of two specific subtasks: (1) Disease named entity recognition and normalization (DNER); (2) Chemical-induced diseases (CID) relation extraction. This paper focuses on CID.

Existing research on CDR extraction can be divided into two categories: rule-based methods [5] and machine learning-based [6], [7], [8], [9], [10], [11] methods. Rule-based methods aim at finding and extracting the heuristic rules for CDR extraction. Lowe et al. [5] develop a simple pattern-based system which could find chemical-induced disease relations within the same sentence. When no patterns matched a document, a heuristic rule is applied to find likely relations. All chemicals in the title, or failing that the first most commonly mentioned chemical in the document, are associated with all diseases in the entire document. Rule-based methods are simple and effective, and have achieved good performance in CDR extraction. But these rules are difficult to apply to a new dataset.

As for machine learning-based relation extraction, feature-based [6], [7], [8], [9] and neural network-based [10], [11] methods are widely used. Feature-based methods focus on designing effective features including lexical and syntactic information. Gu et al. [6] utilize rich linguistic information including various lexical and flat syntactic features for CID task. Zhou et al. [9] extract structured syntactic features based on the shortest dependency path (SDP) between the chemical and disease entities, which are proven effective for CDR extraction. Feature-based methods achieve better performance than rule-based methods by the distributional syntactic-semantic features, however, designing and extracting these features is very time-consuming and laborious.

With the development of neural networks, some studies begin to explore deep contextual semantic representations for relation extraction. Zhou et al. [10] simply adopt a long short-term memory (LSTM) model [12] and a convolutional neural network (CNN) model [13] to get context representations of surface sequences, and have achieved success in CDR extraction. Gu et al. [11] apply CNN to learn context and dependency representations for CDR extraction.

At the same time, many Large-scale knowledge bases (KBs) have been constructed. KBs contain huge amounts of structured data as the form of triples (head entity, relation, tail entity) (denoted as $(h, r, t)$), where relation indicates the relationship between the two entities. These triples could provide rich prior knowledge indicating relations between entities, which are very

Manuscript submitted 21 Sep. 2017. "This work is supported by Natural Science Foundation of China (No.61772109, No.61272375) and the Ministry of education of Humanities and Social Science project (No. 17YJA740076)."

Huiwei Zhou, Yunlong Yang, Shixian Ning, Zhuang Liu, Chengkun Lang, Yingyu Lin and Degen Huang are with the School of Computer Science and Technology at Dalian University of Technology, Dalian, China

(e-mail: zhouhuiwei@dlut.edu.cn, {SDyyl_1949,ningshixian, zhuangliu1992,kunkun}@mail.dlut.edu.cn, lyydut@sina.com, huangdg@dlut.edu.cn).

important for relation extraction. However, a great deal of prior knowledge contained in KBs has not yet been well investigated and utilized. Xu et al. [7] and Pons et al. [8] introduce knowledge features derived from KBs for CDR extraction. Although the performance of CDR extraction has been improved, these methods describe knowledge features as one-hot representations, which assumes that all objects are independent from each other and do not assume the similarities or correlations among objects. Thus one-hot representations cannot take the semantic information into consideration and is easily plagued by dimensionality disaster [14]. For example, the trigger words "*induced*" and "*caused*" both indicate the similar meaning. However, in one-hot representations, the two words are completely different.

To solve these issues, knowledge representation (KR) learning methods are adopted to encode knowledge triple with low-dimensional embeddings of both entities and relations [15], [16], [17]. Knowledge representation learning aims to project entities and relations into a unified dense, real-valued and lowed-dimensional semantic space. Thus semantic correlations of entities and relations can be efficiently measured. In recent years, many knowledge representation learning methods have been proposed, among which translation-based models are simple and effective with the state-of-the-art performance. TransE [15] is a typical translation-based method, which regards a relation $r$ as a translation from the head entity $h$ to the tail entity $t$ with the $h+r \approx t$ in the embedding space, if the triple $(h, r, t)$ holds. TransE applies well to 1-to-1 relations but has issues for 1-to-N, N-to-1 and N-to-N relations. To solve this issue, various methods such as TransH [16] and TransR [17] etc. are proposed. TransH enables an entity to have distinct representations by introducing the mechanism of projecting to the relation-specific hyperplane. That is to say, it positions the relation-specific representation in the relation-specific hyperplane rather than in the same space of entity representations. While TransR builds entity and relation representations in separate entity space and relation-specific spaces, it projects entities from entity space to corresponding relation space and then learn representations via translations between projected entities. Existing knowledge representation learning methods have been widely used to extract general entity relations [15], [16], [17]. However, knowledge representation learning has not yet been explored in the biomedical entity relation extraction.

This paper aims at applying KRs for CDR extraction and investigating the effectiveness of context representations and KRs in biomedical text mining. For a pair of entities in a document, an attention mechanism is employed to select informative context word representations according to their relation representations learned from KBs. Experiments show that both knowledge representations and context representations are effective in CDR extraction.

The major contributions of this paper are summarized as follows:

- We apply knowledge representations learned from KBs to CDR extraction. To the best of our knowledge, this is the first time to evaluate the effectiveness of knowledge representations in biomedical entity relation extraction.
- We propose a neural network-based attention model (NAM) which uses attention mechanism to calculate the weight of context word according to relation representations learned from KBs. It is proved that NAM could effectively combine knowledge and context representations.
- Compared with state-of-the-art systems, our approach could achieve comparable results without any massive handcrafted features and complicated linguistic analysis, thus reduce the time cost and error propagation.

## II. METHODS

The method to CDR extraction can be divided into 5 steps as follows:

(1) Construct relation instance by several heuristic rules for both intra- and inter- sentence level.
(2) Pre-train entity and word representations together.
(3) Extract triples from CDR dataset and CTD, then use them to learn knowledge representation.
(4) Combine context representations with knowledge representations for CDR extraction.
(5) Merge the results of intra- and inter- sentence level to get the final document level results.
(6) Adopt some post-processing rules to further improve the performance.

### A. Relation Instance Construction

Firstly, relation instances for training and testing should be constructed. The instances generated from chemical and disease mentions in the same document are pooled into two groups at intra- and inter-sentence levels, respectively. The former means a chemical-disease mention pair is from the same sentence. The latter means a mention pair is from the different sentences in a document. And if the relation between the two entities of the mention pair is annotated as true in the document, we would take this mention pair as a positive instance; otherwise, we would take this mention pair as a negative instance.

To better understand our extraction rules, take the following sentences from a document (PMID: 12084448) into consideration:

- *S1. **Ifosfamide** (Chemical: D007069) **encephalopathy** (Disease: D001927) presenting with **asterixis** (Disease: D020820).*
- *S2. CNS toxic effects of the antineoplastic agent **ifosfamide** (Chemical: D007069) are frequent and include a variety of neurological symptoms that can limit drug use.*
- *S3. We report a case of a 51-year-old man who developed severe, disabling negative **myoclonus** (Disease: D009207) of the upper and lower extremities after the infusion of **ifosfamide** (Chemical: D007069) for **plasmacytoma** (Disease: D010954).*
- *S4. He was awake, revealed no changes of mental status and at rest there were no further motor symptoms.*
- *S5. Cranial magnetic resonance imaging and extensive*

*laboratory studies failed to reveal **structural lesions of the brain** (Disease: D001927) and **metabolic abnormalities** (Disease: D008659).*

Among all the above sentences, the texts in bold are mentions of chemical or disease entities. Since there are multiple variants of chemical and disease entities, the Medical Subject Headings concept identifiers (MeSH ID) [18] are used to identify chemicals and diseases instead of the entity mentions themselves. The different mentions of the entity which have the same MeSH ID are regarded as the same entity. To read easily, we mark the entity type and MeSH ID in the sentences. Any mentions that occur in parentheses are removed from the sentences.

In the above sentences, the chemical D007069 has intra-sentence level co-occurrence with diseases D001927 in sentences *S1* while it also has intra-sentence level co-occurrence with disease D009207 in sentence *S3*. Moreover, chemical D007069 has inter-sentence level co-occurrence with disease D009207 and D008659 etc. However, not all occurrences of chemicals and diseases are considered as a true CID relation. In this document, only the chemical-disease pairs: D007069-D009207 and D007069-D001927 are labeled as true CID relation. Others are considered as negative instances.

Several heuristic rules are applied to the training and testing datasets for both intra- and inter- sentence level instances. The details are as follows:

*1) Candidate Relation Instance Construction for Intra-Sentence Level*

All chemical-disease pairs that occur in the same sentences are generated as intra-sentence level instances.

For instance, there are three mentions in sentence *S3*. Chemical D007069 and disease D009207 will constitute an intra-sentence level positive instance, while chemical D007069 and disease D010954 will constitute an intra-sentence level negative instance.

*2) Candidate Relation Instance Construction for Inter-Sentence Level*

From the above sentences, we can see that there are a large number of inter-sentence level candidate instances. However, only a few of them are positive. Introducing too many instances would increase the computation load and reduce the performance. Following Gu et al. [6], [11], some heuristic rules are applied to construct the inter-sentence level instances. Although very simple, these rules are quite effective.

(1) Only the chemical-disease entity pairs that are not involved in any intra-sentence level are considered as inter-sentence level instances.

(2) The sentence distance between two mentions in an instance should be less than 3.

(3) If there are multiple mentions that refer to the same entity, we choose the chemical and disease mentions in the nearest distance.

Follow our heuristic rules, chemical D007069 in sentence *S3* and disease D008659 in sentence *S5* form an inter-sentence level instance. However, chemical D007069 in sentence *S1* and disease D008659 in sentence *S5* will be omitted since the sentence distance is more than 3.

In addition, chemical D007069 in sentence *S2* and disease D010954 in sentence *S3* are not considered as an inter-sentence level instance because chemical D007069 already has intra-sentence level instance in sentence *S3* with disease D010954.

*3) Input Sequence Generation*

After getting the candidate relation instances, we generate the input sequence for our NAM as follows:

For intra-sentence level instances, we directly extract the words between chemical-disease pair and expansion of three words on both sides of chemical-disease pair as input word sequence.

For inter-sentence level instances, we concatenate the two sentences where the entity pairs are located and treat it as a sentence, and then extract the input sequence in the same way as intra-sentence level instances.

*B. Entity and Word Representations Pre-training*

Given a candidate input sequence, we need to convert each word or entity in the sequence into a low-dimensional vector. As we use the MeSH ID to represent the entity, the MeSH ID of entity is regarded as a special "word". Then Word2Vec tool[1][19] is applied to pre-train entity and word representations together on the PubMed articles provided by Wei et al. [20], in which chemical and disease entities are recognized and tagged automatically with their corresponding MeSH ID by PubTator[2] tool. The total articles consist of 27 million documents, 185.7 million sentences, and 4.2 million distinct words.

The pre-trained entity representations are used as the initial entity representations for TransE training.

*C. Knowledge Representation Learning*

*1) Triple extraction*

We learn knowledge representations based on the triples extracted from CTD[3] (update January 5, 2017. version: 14906) and CDR dataset [4].

Firstly, all the candidate chemical-disease pairs are extracted according to their MeSH ID from the CDR dataset (all positive and negative instances in training, development and test dataset generated in II.A. *Relation Instance Construction* section) and CTD. Then we extract the relation of these chemical-disease pairs according to CTD. There are three kinds of relations in CTD: *inferred-association*, *therapeutic*, *marker/mechanism*. However, there certainly will be such a situation in which the relation of some chemical-disease pairs extracted from CDR cannot be found in CTD. Therefore, an artificially introduced "*null*" relation is used to complete the relational triples, just like ($e_1$, *null*, $e_2$). Finally, we can get three kinds of relations in CTD: *inferred-association*, *therapeutic*, *marker/mechanism*, and one artificially introduced relation: *null* of 14159 distinct MeSH ID's chemicals, 5714 distinct MeSH ID's diseases and around 1 million CID triples. According to the guidelines of CDR corpus [21], the CID relations in CDR corpus refer to two types of relationship in CTD: "putative mechanistic" relation and "biomarker" relation. In CTD, the two relationships are marked as "*marker/mechanism*". The other relations such as "*therapeutic*", "*inferred-association*" are not annotated in CDR corpus. When learning the relation representation, we use four relations: *inferred-association*, *therapeutic*, *marker/mechanism*, and *null*. These four relations are not the instance relation labeled by CDR dataset. Since we do not use the label of the test dataset to get the relation representation, our method is reasonable and dependable.

---

[1] https://code.google.com/p/word2vec/
[2] http://www.ncbi.nlm.nih.gov/CBBresearch/Lu/Demo/PubTator/.
[3] http://ctdbase.org/

Taking the following example extracted from CTD to explain, the pair of chemical "*doxorubicin* (MeSH ID: D004317)" and the disease "*cardiomegaly* (MeSH ID: D006332)" is annotated with "*marker/mechanism*" in CTD. The chemical, disease and their relation can be represented as a triple (D004317, marker/mechanism, D006332). More generally, we can formalize this triple as $(e_c, r, e_d)$, where $e_c, r, e_d$ indicate a *chemical* entity, a relation, and a *disease* entity respectively.

*2) TransE for knowledge representations learning*

In this paper, with simplicity and good performance in mind, TransE is selected to learn knowledge representations. TransH and TransR are also investigated in the experiments. The basic idea of TransE is illustrated in Fig. 1. TransE could learn the structure information from all the generated triples and encode the chemical representations $e_c$, disease representations $e_d$ and relation representations $\mathbf{r}$ into the continuous vector space $\mathbb{R}^k$. The loss function of TransE is defined as:

$$L = \sum_{(e_c,r,e_d)\in S} \sum_{(e'_c,r,e'_d)\in S'} \max(0, \gamma - \|\mathbf{e}_c + \mathbf{r} - \mathbf{e}_d\| + \|\mathbf{e}'_c + \mathbf{r} - \mathbf{e}'_d\|) \quad (1)$$

where $\gamma > 0$ is a margin between correct triples and incorrect triples, $S$ is the set of correct triples and $S'$ is the set of incorrect triples. CTD only contains correct triples. By convention, these correct triples $(e_c, r, e_d) \in S$ are corrupted by replacing the chemical or disease entity to general the negative triples $(e'_c, r, e_d)$ or $(e_c, r, e'_d)$. When corrupting triple, we follow Wang et al. [16] and assign different probabilities for chemical/disease entity replacement. For those 1-to-N, N-to-1 and N-to-N relations, the "one" side is given more chance to replace.

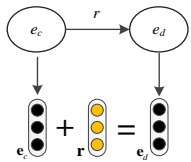

Fig.1. Simple illustration of TransE.

*D. Relation Extraction with Neural Network-based Attention Model*

Knowledge representations learned from KBs are used to extract CDRs along with context representations by NAM. Fig. 2 shows the architecture of our NAM for intra- and inter-sentence level CDR extraction. It consists of three layers: representation layer, attention layer and softmax layer.

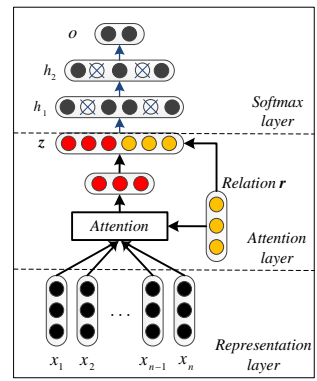

Fig. 2. The architecture of neural network-based attention model.

*1) Representation layer*

Given an input sequence $w = \{w_1, w_2, ..., w_n\}$ with a pair of chemical and disease entities in a document, we map each word into its representation vector to obtain context word representations $x = \{x_1, x_2, ..., x_n\}$, where $x_i \in \mathbb{R}^d$ is a $d$-dimensional word representation. Similarly, the relation $r$ of this chemical-disease pair is also mapped to its representation $\mathbf{r} \in \mathbb{R}^k$ learned through knowledge representation learning.

*2) Attention layer*

The main idea in our attention mechanism comes from Tang et al. [22]. The intuition is that context words do not contribute equally to the semantic meaning of a sequence. Furthermore, the importance of a word should be different if we focus on different relation representation learned from KBs.

Taking the context word representation $x \in \mathbb{R}^{d \times n}$ of an entity pair and their relation representation $\mathbf{r} \in \mathbb{R}^k$ as input, the attention outputs a feature representation $s \in \mathbb{R}^d$. The feature representation $s \in \mathbb{R}^d$ is a weighted sum of the context word representations:

$$s = \sum_{i=1}^{n} \alpha_i x_i \quad (2)$$

where $n$ is the length of context word sequence, $\alpha_i \in [0,1]$ is the weight of $x_i$, and $\sum_{i=1}^{n} \alpha_i = 1$. For each word $x_i$, we use a feed forward neural network to compute its semantic relatedness with the relation of the entity pair. The score function is calculated as follows:

$$g_i = tanh(\mathbf{W}_w(x_i \oplus \mathbf{r}) + b_w) \quad (3)$$

where $\oplus$ denotes the concatenation operation, $\mathbf{W}_w \in \mathbb{R}^{1 \times (d+k)}, b_w \in \mathbb{R}^{1 \times 1}$ are the parameters.

After obtaining $\{g_1, g_2, ..., g_n\}$, the attention weight of each context word can be defined as follows:

$$\alpha_i = \frac{\exp(g_i)}{\sum_{j=1}^{n} \exp(g_j)} \quad (4)$$

In order to make full use of relation representation and better integrate context and knowledge, we concatenate the feature representation $s \in \mathbb{R}^d$ and relation representation $\mathbf{r} \in \mathbb{R}^k$ to get the final attentional representation $z \in \mathbb{R}^{d+k}$:

$$z = s \oplus \mathbf{r} \quad (5)$$

*3) Softmax layer*

The fixed length attentional representation $z \in \mathbb{R}^{d+k}$ is fed into a 2-layer perceptron. We take the non-linear transformation of rectified linear unit (relu) [23] as the activation function. The transformations can be written as follows:

$$h_1 = relu(\mathbf{W}z + b) \quad (6)$$
$$h_2 = relu(\mathbf{W}_{h1}h_1 + b_{h1}) \quad (7)$$

where $\mathbf{W} \in \mathbb{R}^{n_{h1} \times (d+k)}, \mathbf{W}_{h1} \in \mathbb{R}^{n_{h2} \times n_{h1}}, b \in \mathbb{R}^{n_{h1}}, b_{h1} \in \mathbb{R}^{n_{h2}}$ are the parameters.

During the training step, we adopt dropout operation to prevent the over-fitting problem of the hidden units by randomly setting the elements of hidden layers to zero through a proportion $p$. And the feature representations are obtained accordingly:

$$h_1 = dropout(h_1 \odot m_1) \quad (8)$$
$$h_2 = dropout(h_2 \odot m_2) \quad (9)$$

where $\odot$ is an element-wise multiplication and $m_1, m_2$ are the mask embeddings whose elements follow the Bernoulli distribution with the proportion $p$.

Finally, the feature representation $h_2$ is fed into a softmax function to compute the confidence of CDR:

$$o = softmax(\mathbf{W}_o h_2 + b_o) \quad (10)$$

where $o \in \mathbb{R}^{n_o}$ is the output, $\mathbf{W}_o \in \mathbb{R}^{n_o \times n_{h2}}$ is the weight matrix and $b_o \in \mathbb{R}^{n_o}$ is the bias.

*E. Relation Merging*

After relation extraction, the intra- and inter-sentence level extraction results are merged as the final document-level result.

There may be multi-instances for a pair of entities in the document, this may result in inconsistent results for the entity pair [6]. If at least one of these instances is predicted as positive by our model, then we would believe this entity pair has a true CID relation.

*F. Post-processing*

To further improve the performance, we employ some heuristic post-processing rules to identify the missed relations and remove redundant relations. The rules are listed as follows:

*1) Focused chemical rule*

When no CDR extracted by NAM in the document, optionally, Lowe et al. [5] apply a focused chemical rule. For this, they assume that if the chemical occurs in the title of the document, it is in focus. If no chemical in the title, they assume that the first most commonly mentioned chemical in the abstract is in focus. And all focused chemicals are associated with all diseases in the entire abstract. Inspired by Lowe et al. [5], we also apply this heuristic rule to identify the missed relations.

*2) Hypernym filtering rule*

The goal of the CID task is to extract the relationships between the most specific disease and chemical entities. However, there exist hypernym or hyponym relationship between concepts of diseases or chemicals, where a concept was subordinate to another more general concept. The relations between hyponym concepts should be considered. However, the relations between hypernym concepts should be removed.

Therefore, following Gu et al [6], we use the Medical Subject Headings (MeSH)-controlled vocabulary [18] to determine the hypernym relationship between entities in a document. Then we remove these hyper-relation instances that involve entities which are more general than other entities already participating in the candidate instances.

III. EXPERIMENTS AND RESULTS

In this section, we first present a brief introduction of the CDR corpus and our experiments settings, and then systematically evaluate the performance of our approach with the golden entities on the test dataset.

*A. Experiment Setup*

**Dataset.** The CDR corpus contains a total of 1500 PubMed articles: 500 each for the training, development and test set. Table I shows the statistics on the number of CID relations for the three datasets.

TABLE I
STATISTICS OF THE CDR DATASET

| Task dataset | No. of Articles | No. of CID relations |
|---|---|---|
| Training | 500 | 1038 |
| Development | 500 | 1012 |
| Test | 500 | 1066 |

We combine the training set and development set as a new training set for training NAM. In our paper, we use the golden standard annotated entities provided by BioCreative V to evaluate our relation extraction system. The golden standard annotated entities imply that both the disease and chemical entities have been correctly labeled. And all the results of state-of-the-art systems reported in our paper are evaluated with the gold standard annotations. Therefore, it is very fair and reasonable to compare these results in which it could avoid the influence of the NER tools. The evaluation is reported by official evaluation toolkit[4], which adopts Precision (*P*), Recall (*R*) and F-score (*F*) to measure the performance.

**Experiment Settings.** The set of parameters that produce the best results based on 10-fold cross-validation on the training set are chosen for our experiments. The dimensions of word, entity and relation representations are set to 100 for the consideration of computational complexity. Note that for those words and entities that do not occur in the pre-training corpus (PubMed articles [20]), we take a random embedding with the uniform distribution in $[-0.25, 0.25]$ to initialize them. The dimensions of 2-layer perceptron in softmax layer are $\{100, 50\}$ with the dropout proportion $p = 0.5$. The NAM is trained by AdaGrad technique [24] with a learning rate 0.01 and a mini-batch size of 32. In addition, we implement TransE using the code[5] provided by Lin et al. [17], and apply default settings: learning rate $\lambda = 0.001$, margin $\gamma = 1$, etc. Our model is implemented with an open-source deep learning library Keras [25] and is publicly available at https://github.com/Xls1994/CDRextracion.

---

[4] http://www.biocreative.org/tasks/biocreative-v/track-3-cdr/

[5] https://github.com/thunlp/KB2E

*B. Comparison of Baseline Methods*

We compare our NAM with several baseline methods, including TransE, CNN, LSTM, and BiLSTM.

- **TransE:** This is a naive method of relation extraction with KBs. Given a candidate entity pair $(e_c, e_d)$, we need to infer whether $(e_c, e_d)$ has a true CID relation. For each candidate entity pair, we calculate the cosine similarity score between the difference vector $\mathbf{v}_r = \mathbf{e}_c - \mathbf{e}_d$ and four different candidate relation representations $\mathbf{r}$ learned by TransE respectively. According to the guidelines of CDR corpus [21], the CID relations in CDR corpus refer to the relationship "*marker/mechanism*" in CTD. We would believe that these entity pairs which have the maximum similarity with the relation "*marker/mechanism*" have the true CID relation.
- **CNN:** This method applies to CNN with convolution, max pooling operation. In CNN, 100 feature maps are learned for each of four different filter sizes $\{1, 2, 3, 4\}$.
- **LSTM:** This method uses the standard LSTM proposed by Hochreiter and Schmidhuber [12]. The dimension of hidden layer in LSTM is 100 and the last hidden vector in hidden layer is used as context representations for classification.
- **BiLSTM:** This method uses the bi-directional LSTM (BiLSTM) model for CDR extraction. Specifically, the last hidden vectors in both directions of LSTM are concatenated as context representations for classification.

Experimental results are shown in Table II. From the table, we can see that:

(1) **TransE** only uses pure prior knowledge and lacks effective contextual information, which leads to poor performance.

(2) **CNN**, **LSTM** and **BiLSTM** use context information and achieve a better performance than **TransE**, which suggests that context information is effective for CDR extraction. **CNN** achieves a slightly improvement compared with **LSTM** and **BiLSTM** due to the fact that local context features extracted from CNN are more effective than long-term sequence features extracted from LSTM.

(3) By incorporating attention mechanism, **NAM** could effectively fuse knowledge and context representations and significantly outperform all baseline methods. It indicates that context and knowledge information are both beneficial to CDR extraction.

Note that the recall of CDR merging (document level) is the sum of recalls of intra-sentence level and inter-sentence level because only entity pairs which are not covered by intra-sentence level are considered as inter-sentence level instances. That is to say, these two level instances are totally irrelevant and completely separated, so the recall of document level is the sum of intra- and inter- sentence level.

*C. Effects of Attention Mechanism*

In order to explore the effects of the attention mechanism, we further study some variants of our model according to different combinations of relation representation and context representation in Formula (3) and (5).

- **CN-CN:** This method only uses context representations (CN) to get the attention weights by: $g_i = tanh(\mathbf{W}_a x_i + b_a)$, and gets the output of attention layer without concatenating knowledge representations (KN): $z = \sum_{i=1}^{n} \alpha_i x_i$.
- **KNCN-CN:** This method uses both knowledge representations and context representations to get the attention weight by: $g_i = tanh(\mathbf{W}_w (x_i \oplus \mathbf{r}) + b_w)$, but gets the output of attention layer without concatenating knowledge representation: $z = \sum_{i=1}^{n} \alpha_i x_i$.
- **CN-KNCN:** This method only uses context representations to get the attention weights by $g_i = tanh(\mathbf{W}_a x_i + b_a)$, but gets the output of attention layer with both context representations and knowledge representations by $z = s \oplus \mathbf{r}$.

Table III shows the results of different variants models. From the Table we can see that:

(1) The performance of **KNCN-CN** and **CN-KNCN** is significantly higher than **CN-CN** due to the fact that **KNCN-CN** and **CN-KNCN** extra use the knowledge representations obtained from KBs through TransE model. Knowledge representations could efficiently encode relational knowledge in a low-dimensional space and serve as an indicator of the entity's relationship, thus significantly improve the performance of CDR extraction.

(2) **NAM** achieves the best performance in all methods. Compared with other methods, knowledge representations are integrated into the **NAM** at the attention level and the classification level respectively. Thus **NAM** could better integrate context information and prior knowledge through the proposed attention mechanism.

*D. Effects of Knowledge Representation Learning Methods*

We further investigate several knowledge representation learning methods, including: **TransE**, **TransH**, and **TransR**. **NAM** with (**Random**, **TransE**, **TransH**, **TransR**) means that NAM uses different knowledge representation learning method for CDR extraction. **Random** means the relation representations are randomly initialized with the uniform distribution in $[-0.25, 0.25]$ and fine-tuned during training phase. Both **TransH** and **TransR** use the same parameters as **TransE** and train 500 epochs.

Experimental results are shown in Table IV, from which we can see that:

(1) **NAM with TransE/TransH/TransR** perform the best compared with **NAM with Random** at both intra- and inter-sentence levels, which indicates that knowledge representations could reveal semantic correlations of entities and relations and provide more exact information than random initialization.

(2) **NAM with TransE** performs the best compared with **NAM with TransH/TransR**. TransH models the relations as translating operations on relation-specific hyperplanes, allowing entities to have different representations when involved in different relations. So different relation representations are on the distinct hyperplanes. Similarly, TransR builds entity and relation representations in separate entity space and multiple relation-specific spaces, and performs translation in the corresponding relation spaces. That is to say, in TransR, different relation representations are on the distinct spaces. However, TransE simply models entities and relations in a union space, which is the same as our hypothesis that models the entities and relations in the same space. Therefore, TransE may be more suitable for our model.

*E. Effects of the Pre-trained Word Representations*

In this section, we explore the effect of several different pre-trained 100-dimensional word embeddings based on the NAM, including Random, GloVe27B [26], GloVe6B [26], W2V100B [19] and PubMed [20].

- Random means all the word embeddings are initialized with the uniform distribution in $[-0.25, 0.25]$.
- GloVe27B is the pre-trained GloVe embedding[6] on Twitter, which contains 27B tokens, 1.2M vocab.
- GloVe6B is the pre-trained GloVe embedding on Wikipedia 2014 and Gigaword 5, which contains 6B tokens, 400K vocab.
- W2V100B is the pre-trained word embedding on Google News, which contains 100B tokens, and 3B vocab. The dimension of W2V100B is 300 since Google only provides 300-dimensional publicly word embedding trained on Google News[7].
- PubMed is the word embedding actually used in our model, which is trained on all the PubMed articles [20] using Word2Vec [19] tool.

According to our statistical results, a total of 41.09% words in the dataset are not found in GloVe27B; and 18.63% words are not found in Glove6B and 36.26% words are not found in W2V100B and 1.34% words are not found in PubMed. Here, though GloVe27B has a larger number of tokens and vocab than GloVe6B, GloVe6B covers more words in the CDR dataset for the reason that GloVe27B is trained on casual Twitter corpus while GloVe6B is trained on Wikipedia, which is more formal and covers a wider area. Noting that, to solve the problem of unknown words, we initialize them from uniform distribution in $[-0.25, 0.25]$. Note that the relation representations both use the same embeddings learned by TransE.

Fig. 3 shows the document level results with different word embedding. From Fig. 3, we can see that the pre-trained word embedding on PubMed articles significantly outperform the other word embedding and yield a 4.52% improvement compared with Random, a 3.37% improvement compared with GloVe27B, a 2.36% improvement compared with GloVe6B and a 1.28% improvement compared with W2V100B. These show that the pre-trained word embeddings on PubMed articles contain more relevant domain-specific semantic information than other pre-trained word embeddings, which results in a good CDR performance.

TABLE II
COMPARISON WITH BASELINE METHODS.

| Method | Intra-sentence level | | | Inter-sentence level | | | Relation merging | | |
|---|---|---|---|---|---|---|---|---|---|
| | P (%) | R (%) | F (%) | P (%) | R (%) | F (%) | P (%) | R (%) | F (%) |
| **TransE** | 43.83 | 32.00 | 37.00 | 19.86 | 13.79 | 16.28 | 32.15 | 45.79 | 37.78 |
| **CNN** | 46.40 | 51.31 | 48.73 | 32.26 | 3.75 | 6.72 | 45.05 | 55.06 | 49.56 |
| **LSTM** | 50.46 | 45.87 | 48.06 | 24.42 | 3.94 | 6.79 | 46.54 | 49.81 | 48.12 |
| **BiLSTM** | 49.16 | 49.34 | 49.25 | 23.45 | 6.37 | 10.03 | 43.68 | 55.72 | 48.97 |
| **NAM** | 63.47 | 60.32 | 61.86 | 55.93 | 12.38 | 20.28 | 62.05 | 72.70 | 66.95 |

TABLE III
EFFECTS OF ATTENTION MECHANISM.

| Method | Intra-sentence level | | | Inter-sentence level | | | Relation merging | | |
|---|---|---|---|---|---|---|---|---|---|
| | P (%) | R (%) | F (%) | P (%) | R (%) | F (%) | P (%) | R (%) | F (%) |
| **CN-CN** | 49.44 | 50.28 | 49.86 | 28.74 | 4.50 | 7.78 | 46.67 | 54.78 | 50.41 |
| **KNCN-CN** | 61.45 | 60.41 | 60.92 | 48.68 | 8.63 | 14.66 | 59.50 | 69.04 | 63.92 |
| **CN-KNCN** | 61.88 | 60.60 | 61.23 | 56.95 | 11.91 | 19.71 | 61.01 | 72.51 | 66.27 |
| **NAM** | 63.47 | 60.32 | 61.86 | 55.93 | 12.38 | 20.28 | 62.05 | 72.70 | 66.95 |

TABLE IV
EFFECTS OF KNOWLEDGE REPRESENTATION LEARNING

| Method | Intra-sentence level | | | Inter-sentence level | | | Relation merging | | |
|---|---|---|---|---|---|---|---|---|---|
| | P (%) | R (%) | F (%) | P (%) | R (%) | F (%) | P (%) | R (%) | F (%) |
| **TransE** | 43.83 | 32.00 | 37.00 | 19.86 | 13.79 | 16.28 | 32.15 | 45.79 | 37.78 |
| **TransH** | 38.03 | 41.75 | 39.80 | 15.77 | 17.07 | 16.40 | 26.98 | 58.82 | 37.00 |
| **TransR** | 33.48 | 44.00 | 38.02 | 14.53 | 16.14 | 15.29 | 24.80 | 60.13 | 35.12 |
| **NAM with Random** | 64.22 | 58.26 | 61.09 | 53.80 | 12.01 | 19.63 | 62.16 | 70.27 | 65.96 |
| **NAM with TransE** | 63.47 | 60.32 | 61.86 | 55.93 | 12.38 | 20.28 | 62.05 | 72.70 | 66.95 |
| **NAM with TransH** | 62.71 | 60.41 | 61.54 | 54.79 | 11.26 | 18.68 | 61.31 | 71.67 | 66.09 |
| **NAM with TransR** | 65.14 | 58.72 | 61.77 | 55.30 | 11.25 | 18.70 | 63.33 | 69.98 | 66.48 |

[6] https://nlp.stanford.edu/projects/glove/

[7] https://code.google.com/archive/p/word2vec/

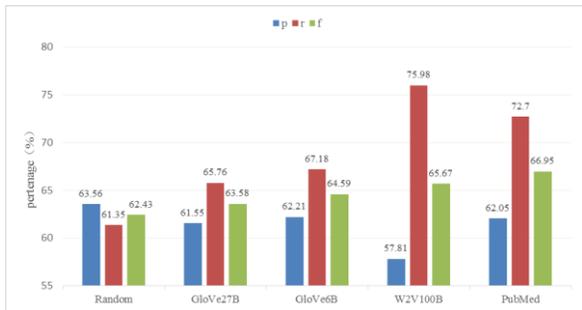

Fig. 3. Document-level result of NAM with different word embedding.

### F. Effects of Post-processing

In this section, we explore the effect of the post-processing rules to the document level results of **NAM** one by one. The results of post-processing are shown in Table V.

From Table V, we can see that the performance of CDR extraction is improved by 0.1% F-score when the focused rule is added. This rule is able to pick the most likely CDR back and improve the recall significantly with a slight decrease in the precision. After the addition of hypernym filtering rule, the performance has been further improved and reached 67.94% F-score. The hypernym filtering rule improves the precision of our model by removing some of the false positives from all positive predictions. As a supplement to the system, post-processing has a very strong effectiveness on the CDR extraction.

TABLE V
RESULT OF THE POST-PROCESSING.

| Method | P (%) | R (%) | F (%) |
|---|---|---|---|
| **NAM** | 62.05 | 72.70 | 66.95 |
| **NAM**+ Focused chemical rule | 59.08 | 77.49 | 67.05 |
| **NAM**+ Focused chemical rule +Hypernym filtering rule | 62.06 | 75.05 | 67.94 |

## IV. DISCUSSION

### A. Comparison with Related Work

We compare our NAM with some related systems of BioCreative V CDR task in Table VI. All the systems are reported on the test dataset with golden standard entity annotations. We mainly divide these different methods into three groups: rule-based methods, Machine Learning-based methods without additional resources (ML without KBs), and ML methods using external knowledge bases (ML with KBs).

TABLE VI
COMPARISON WITH RELATED WORK.

| Method | System | P (%) | R (%) | F (%) |
|---|---|---|---|---|
| Rule-based | Lowe et al. [5] | 59.29 | 62.29 | 60.75 |
| ML without KBs | Gu et al. [6] | 62.00 | 55.10 | 58.30 |
| | Zhou et al. [10] | 55.56 | 68.39 | 61.31 |
| | Gu et al [11] | 55.70 | 68.10 | 61.30 |
| ML with KBs | Xu et al. [7] | 65.80 | 68.57 | 67.16 |
| | Pons et al. [8] | 73.10 | 67.60 | 70.20 |
| | Peng et al. [27] | 68.15 | 66.04 | 67.08 |
| | Ours | 62.06 | 75.05 | 67.94 |

Compared the results of the different methods in Table VI, ML with KBs could significantly outperforms the methods without the help of KBs (Rule-based method and ML without KBs).

In ML methods with KBs, Xu et al. [7] use four free available large-scale prior knowledge bases to extract the prior knowledge features, which contributes 16.43% F-score to CDR extraction performance. Besides some commonly used freely available KBs, such as UniProt, CTD and UMLS etc., the commercial system, Euretos Knowledge Platform, is also used in Pons et al. [8], which leads to the best performance with an F-score of 70.20%. Peng et al. [27] extract one-hot knowledge features based on CTD and MeSH databases and achieves an F-score of 67.08%. Compared with other ML with KBs methods, including Xu et al. [7] (67.16% F-score), Pons et al. [8] (70.20% F-score) and Peng et al. [27] (67.08% F-score), the main difference of NAM is that our method uses the proposed attention mechanism to combine the knowledge representations obtained from TransE and context representations. This will enable our model to efficiently compute semantic links between entities and relationships in low-dimensional space, resulting in an increase in CDR extraction performance. In addition, we do not need extensive manual feature engineering and our method would be more universal and easier to apply.

### B. Statistical Significance of Different Methods

To see whether our method yields significant difference, *t*-test statistics is conducted by 10-fold cross validation on the training and development datasets. The average *F*-score improvement of method 1 compared to method 2 and *P*-values is shown in Table VII.

TABLE VII
STATISTICAL SIGNIFICANCE OF PERFORMANCE OVER 10-FOLD CROSS VALIDATION.

| Method 1 | Method 2 | Average F-score improvement (%) | P-values |
|---|---|---|---|
| **NAM with Random** | CN-CN | 10.73 | 2.46E-05 |
| **NAM with TransE** | NAM with Random | 2.12 | 0.45E-02 |
| **NAM with TransE** | CN-CN | 12.84 | 6.31E-07 |
| **NAM with TransE** | KNCN-CN | 1.33 | 0.29E-1 |
| **NAM with TransE** | CN-KNCN | 0.67 | 0.11 |

From Table VII, we can see *t*-test for **NAM with Random** vs. **CN-CN** results a *P*-value of 2.46E-05, which shows a significant difference with knowledge representations introduced. Furthermore, the difference between **NAM with TransE** and **NAM with Random** is also significant (*P*-value<0.05), which indicates that learning relation representations by TransE outperforms random initialization significantly. And statistical analysis also shows significant difference between **NAM with TransE** and **CN-CN**. **NAM with TransE** also shows statistically significant improvements in comparison to **KNCN-CN** (P-value<0.05), demonstrating the effectiveness of concatenating relation representation with the context representations.

### C. Visualization of Attention

To understand our attention mechanism clearly, we visualize the attention weights of two example sequences in the form of heat map in Fig. 4. The darker the color, the higher the attention weight. All the words are converted to lowercase. The entities are converted to their corresponding MeSH ID. For the first sequence "background: *acetaminophen* (*Chemical: D000082*)

*induced hepatotoxicity (Disease: D056486) is the most common cause of acute liver failure (Disease: D017114) in the uk*". In Fig. 4, the trigger words "*cause*" and "*induced*" have the higher weight score than other words, when paying attention to relation representations "*marker/mechanism*". For the second sequence "*the fda showed **clarithromycin** (Chemical: D017291) and **ciprofloxacin** (Chemical: D002939) to be the most frequently associated with the development of **mania** (Disease: D001714)*". The trigger word "*associated*" has the higher weight score than other words. Therefore, we believe that the NAM model could identify the important contextual word effectively.

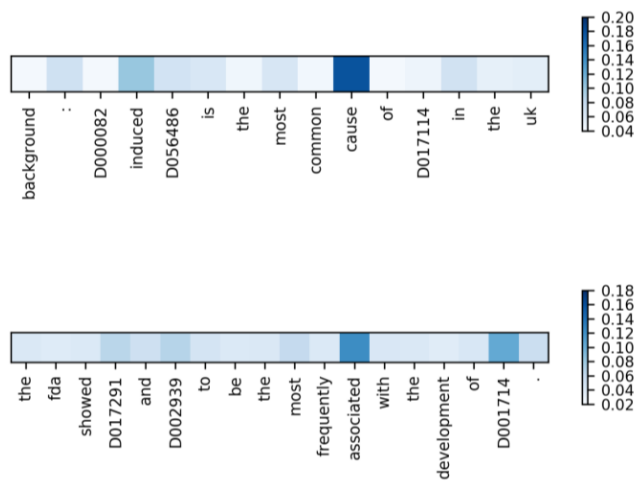

Fig. 4. Visualization of attention weight by heat map.

### D. Error Analysis

We perform an error analysis on the output of our final results to detect the origins of false positives (FP) and false negatives (FN), which are categorized in Fig. 5. We list some examples wrongly extracted by our model to better understand our results.

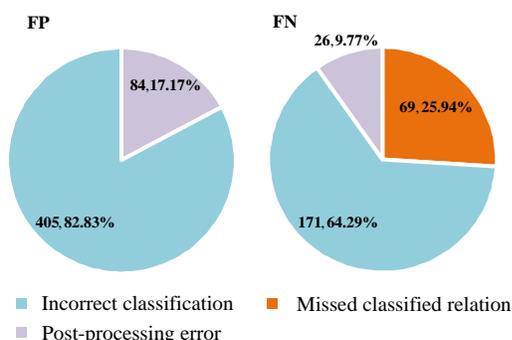

Fig. 5. Origins of FP and FN errors.

For FP in Fig. 5, two main error types are listed as follows:

- Incorrect classification: In spite of the detailed semantic representations, 405 FP errors come from the incorrect classification made by our model. Among the 405 FP errors, 313 FP come from the intra-sentence level and 92 FP come from the inter-sentence level. For the sentence "*Baseline electrocardiogram abnormalities and market elevations not associated with **myocardial necrosis** (Disease: D009202) make accurate diagnosis of **myocardial infarction** (Disease: D009203) difficult in patients with **cocaine** (Chemical: D003042)-associated **chest pain** (Disease: D003042). (PMID: 12359538)*" in the test set, the pair of chemical *cocaine* (D003042) and disease *myocardial infarction* (D009203) is annotated as CID relation, while the pair of chemical *cocaine* (D003042) and disease *myocardial necrosis* (D009202) is not. However, our model wrongly extracted the pair of *cocaine* (D003042) and *myocardial necrosis* (D009202). The two diseases have similar context and it is hard for our model to find their difference from the context.

- Post-processing error: The focused rules bring 84 false CDR, with a proportion of 17.17%. For example, there is no CDR found by our model in the document of the title "*Induction of rosaceiform **dermatitis** (Disease: D003872) during treatment of **facial inflammatory dermatoses** (Disease: D005148) with **tacrolimus** (Chemical: D016559) ointment. (PMID: 15096374)*". Following the focused chemical rule, the chemical *tacrolimus* (D016559) in the title is associated with the disease *dermatitis* (D003872) and *facial inflammatory dermatoses* (D005148) in the abstract. However, the two relations are not the true CID relations.

For FN in Fig. 5, three main error types are listed as follows:

- Post-processing error: The hypernym filter rule removes 26 real CDR, with a proportion of 9.77%. For the sentence "*These complications have included clinical deterioration and intracranial vascular **thrombosis** (Disease: D013927) in patients with SAH, arteriolar and capillary fibrin **thrombi** (Disease: D013927) in patients with fibrinolytic syndromes treated with **EACA** (Chemical: D015115), or other **thromboembolic phenomena** (Disease: D013923). (PMID: 448423)*", the pair of chemical *EACA* (D015115) and disease *thrombosis/thrombi* (D013927) and the pair of chemical *EACA* (D015115) and disease *thromboembolic phenomena* (D013923) are extracted by our model. The hypernym filter rule removes the relation of chemical *EACA* (D015115) and disease *thrombosis/thrombi* (D013927) because *thrombosis/thrombi* (D013927) is the hypernym of *thromboembolic phenomena* (D013923). However, the CID relation pair of the chemical *EACA* (D015115) and disease *thrombosis/thrombi* (D013927) is annotated as CID relation in the test set.

- Missed classified relation: 69 inter-sentence level instances are removed by the heuristic rules in section II.A *Relation Instance Construction*, which are not classified by our system at all because the sentence distance between these chemical and disease entities are more than 3.

- Incorrect classification: Among the 266 CDR that have not been extracted, our model misclassifies 171 positive cases (43 intra-sentence level positive cases and 128 inter-sentence level positive cases) as negatives due to complex syntactic and latent semantic information of entity pairs. For the sentence "*BACKGROUND: Several studies have demonstrated liposomal **doxorubicin** (Chemical: D004317) to be an active antineoplastic agent in platinum-resistant* *ovarian cancer, with dose limiting toxicity of the standard dosing regimen (50 mg/m(2) q 4 weeks) being severe **erythrodysesthesia** (Disease: D060831) and **stomatitis** (Disease: D013280). (PMID: 10985896)"* in the test set, the pair of chemical **doxorubicin** (*D004317*) and disease **erythrodysesthesia** (*D060831*), and the pair of chemical **doxorubicin** (*D004317*) and disease **stomatitis** (*D013280*) are annotated as CID relations. However, due to the complex syntactic and latent semantic inference, our model fails to extract both CID relations.

## V. Conclusion

In this paper, we introduce the knowledge representations learned from KBs into the CDR extraction task and develop an effective attention mechanism to capture the importance of each context word according to its semantic relatedness with the relation representations. Experimental results on the BioCreative V CDR dataset show that the attention mechanisms can effectively fuse knowledge and context representations, and the performance of CDR extraction has been significantly improved with the help of knowledge representations. The proposed NAM model could be comparable to state-of-the-art CDR systems without any handcrafted features.

This paper only uses a typical chemical-diseases knowledge bases CTD for knowledge representation learning. However, many other biomedical knowledge bases, such as UMLS, MESH, UniProt and the commercial system Euretos Knowledge Platform, etc., have not been used by us yet. The heterogeneity and imbalance of the entities and relations in these knowledge bases are the main problems that restrict knowledge representation learning. A unified knowledge representation space can be established to project these entities and relations from different sources into the same semantic space. How to use it in biomedical entity extraction is still a challenging task. In the future, we would like to explore richer knowledge information to enhance the performance of CDR extraction.

ACKNOWLEDGMENT

This work is supported by Natural Science Foundation of China (No.61772109, No.61272375) and the Ministry of education of Humanities and Social Science project (No. 17YJA740076). Huiwei Zhou is the corresponding author of this paper.